\documentclass[twocolumn]{article}

\usepackage{algorithm}
\usepackage{algorithmic}
\usepackage{fullpage,times}
\usepackage{parskip}
\usepackage{multirow}
\usepackage{wrapfig}
\usepackage{authblk}
\usepackage{lipsum}  
\makeatletter
\newcommand{\printfnsymbol}[1]{%
  \textsuperscript{\@fnsymbol{#1}}%
}
\makeatother
\makeatletter
\def\hlinewd#1{%
\noalign{\ifnum0=`}\fi\hrule \@height #1 %
\futurelet\reserved@a\@xhline}
\makeatother

\newcommand{\largerAppendixTitle}[1]{\noindent{\LARGE\bfseries #1}}

\date{}
\newcommand{\z}{z}
\newcommand{\x}{x}
\newcommand{\f}{f}
\usepackage{bm} 
\usepackage{qiangstyle}
\usepackage[bibstyle=nature, citestyle=numeric-comp,sorting=none,doi=false,isbn=false,url=false]{biblatex}
\begin{document}

\definecolor{commentcolor}{RGB}{110,154,155}   
\definecolor{inputcolor}{RGB}{255, 105, 180}   
\newcommand{\PyComment}[1]{\ttfamily\textcolor{commentcolor}{\# #1}}  
\newcommand{\PyInput}[1]{\ttfamily\textcolor{inputcolor}{\# #1}}
\newcommand{\PyCode}[1]{\ttfamily\textcolor{black}{#1}} 

\title{Flow Perturbation to Accelerate Unbiased Sampling of Boltzmann distribution}
\author[1]{Xin Peng} 
\author[1]{Ang Gao\thanks{Corresponding: anggao@bupt.edu.cn}}
\affil[1]{School of Science, Beijing University of Posts and Telecommunications, Beijing 100876}
\maketitle
\abstract
Flow-based generative models have been employed for sampling the Boltzmann distribution, but their application to high-dimensional systems is hindered by the significant computational cost of obtaining the Jacobian of the flow during reweighting. To overcome this challenge, we introduce the flow perturbation method, which incorporates optimized stochastic perturbations into the flow, eliminating the need for Jacobian calculations. Our method achieves unbiased sampling of the Boltzmann distribution with orders of magnitude speedup compared to both brute force Jacobian calculations and the Hutchinson estimator. Notably, it accurately sampled the Chignolin protein with all atomic Cartesian coordinates explicitly represented, which, to our best knowledge, is the largest molecule ever Boltzmann sampled in such detail using generative models.  
\section*{Introduction}

The Boltzmann distribution defines the probability of states of a many-body system at thermal equilibrium.  Sampling this distribution is notably challenging due to the presence of numerous meta-stable states separated by high energy barriers\cite{Frauenfelder91}. Traditional sampling methodologies, such as Molecular Dynamics (MD) \cite{PhysRev.159.98} and Markov Chain Monte Carlo (MCMC)\cite{Metropolis:1953vj}, operate through making small incremental movements within the configuration space, which makes them inefficient in crossing energy barriers. Enhanced sampling techniques, including replica exchange\cite{Hukushima:1996uz, PhysRevLett.57.2607}, umbrella sampling\cite{TORRIE1977187}, metadynamics\cite{Laio:2002wm}, and transition path sampling\cite{10.1063/1.475562}, offer avenues to overcome these barriers, yet they introduce their own set of challenges, such as the computational overhead of simulating numerous replicas\cite{Ballard:2009wa} and the challenge of identifying suitable collective variables\cite{Laio:2002wm, 10.1063/1.475562}.

Recent advancements in deep generative models have opened new avenues for generating configurations of many-body systems\cite{Zheng:2024ww, NEURIPS2022_994545b2, xu2022geodiff, pmlr-v162-hoogeboom22a, wang2023generating, Wang:2024vq, Noe:2019uu, midgley2023se, PhysRevD.100.034515, pmlr-v202-kohler23a}. These models map samples from a simple prior distribution to a complex target distribution through a series of transformations parameterized by deep neural networks. One key advantage of deep generative models is their ability to generate independent samples in parallel, making them less likely to get stuck in meta-stable states compared to traditional methods like MD or MCMC. Recently, deep generative models have demonstrated the ability to generate configurations of complex molecules such as proteins with considerable accuracy, drawing significant attention from the scientific community\cite{Janson:2023wc, Wu:2024vv, Ingraham:2023ts, Watson:2023ut, yim2023fast, Abramson:2024vm}. However, a significant challenge with these models is that the samples they generate typically do not unbiasedly sample the Boltzmann distribution, limiting their ability to accurately capture equilibrium properties that are crucial for applications in computational chemistry and physics.

Normalizing Flows (NFs) \cite{10.5555/3045118.3045281, NEURIPS2018_d139db6a, dinh2017density, dinh2015nice} have emerged as a promising solution to this sampling problem. NFs are a special class of deep generative models that transforms the prior distribution  through a series of bijective and differentiable neural network layers. By reweighting trajectories generated by NFs based on the generalized work they produced, one can achieve unbiased sampling of the Boltzmann distribution \cite{Noe:2019uu, SNF_Wu2020, PhysRevD.100.034515, PhysRevLett.126.032001, PhysRevE.101.023304, Ahmad_2022, Wirnsberger_2022, pmlr-v202-kohler23a, Invernizzi:2022vt, PhysRevResearch.4.L042005, Sbailo:2021vr, Gabrie:2022wf, schonle:hal-04404948, molinataborda2024active, midgley2022flow}.  However, calculating the generalized work requires computing the determinant of the Jacobian of the flow, which necessitates performing \(D\)  backpropagation passes through all the layers of the flow, where \(D\) is the dimensionality of the system. This process is computationally expensive, posing a significant challenge for applying NFs to high-dimensional systems.  Moreover, the Jacobian calculation is also a significant hurdle in the training of NFs.  To mitigate this issue, specialized neural network layers with easily calculable Jacobian, such as NICE\cite{dinh2015nice}, RealNVP\cite{dinh2017density}, and Glow\cite{NEURIPS2018_d139db6a}, are typically used. However, these layers limit the expressivity of the flow, making it challenging for NFs to model highly complex distributions.

Continuous Normalizing Flows (CNFs), a special limiting case of NFs, have recently been applied  to sample the Boltzmann distribution\cite{pmlr-v119-kohler20a, klein2023equivariant, Wang:2024vq}. CNFs use neural Ordinary Differential Equations (neural ODEs)\cite{NEURIPS2018_Chen} to establish the mapping between the prior and target distributions.  A notable benefit of CNFs over traditional NFs is that they are easier to train, thanks to advancements in simulation-free training methods\cite{lipman2023flow, liu2023flow, albergo2023building, tong2023improving, pmlr-v202-pooladian23a, chen2024flow}. This ease of training allows CNFs to more effectively model complex distributions. For example, the ODE formulation\cite{song2021scorebased, NEURIPS2022_a98846e9} of the diffusion model has achieved state-of-the-art performance in generating images\cite{song2021scorebased,song2021denoising, NEURIPS2022_a98846e9} and molecular configurations\cite{NEURIPS2022_994545b2}. Additionally, the optimal-transport flow has been shown to be able to generate high-quality protein backbone configurations\cite{yim2023fast}.

Despite being easier to train, obtaining the Jacobian determinant of a CNF, which is essential for unbiased sampling of the Boltzmann distribution, requires integrating the Jacobian trace of its velocity field along the flow. The Jacobian trace calculation still necessitates $D$ backpropagation passes, making it challenging to apply CNFs to sample the Boltzmann distribution of high-dimensional systems\cite{pmlr-v119-kohler20a}. The Hutchinson estimator\cite{1989AHutch} has been used to reduce this computational cost\cite{Wang:2024vq}. However, it complicates the reweighting process, leading to convergence issues\cite{pmlr-v119-kohler20a, erives2024verletflowsexactlikelihoodintegrators} and systematic biases in the reweighted results\cite{pmlr-v119-kohler20a}. As a result, CNFs have only achieved unbiased sampling for small molecules\cite{klein2023equivariant}.

In this work, we introduce the flow perturbation method, a novel approach designed to accelerate the unbiased sampling of the Boltzmann distribution of high-dimensional systems.  By leveraging the fact that stochastic trajectories, like their deterministic counterparts, can also be reweighted according to their generalized work\cite{SNF_Wu2020, doi:10.1073/pnas.1106094108}, we introduce specific stochastic perturbations into the flow model, which allows us to avoid  Jacobian calculations. Additionally, the magnitude of the perturbations are optimized to minimize fluctuations in the generalized work, ensuring efficient reweighting. 

Our flow perturbation method is not the first attempt to incorporate stochasticity into flow models. Wu et al\cite{SNF_Wu2020} introduced the Stochastic Normalizing Flow (SNF), which inserts stochastic layers between deterministic NF layers. However, their approach still requires calculating the Jacobian of the NF layers, potentially limiting its applicability to high dimensional systems. Other sampling methods based on stochastic differential equations (SDEs) also exist\cite{zhang2022path, vargas2023denoising, zhang2024diffusion}, but numerically solving the SDE requires breaking it down into numerous small steps using Euler-Maruyama scheme, which is computationally expensive to implement. By comparison, our method just appends a stochastic perturbation at the end of the flow, which allows  the flow to be integrated using efficient ODE solvers\cite{ascher1998computer}, thus significantly reducing the computational load. Moreover, the dimensionality of trajectory space  in our method is much lower than in SDE-based approaches, making it easier to explore the trajectory space using techniques like MCMC. These advantages make the flow perturbation method particularly well-suited for sampling the Boltzmann distribution of high-dimensional systems.

Our method has been tested on high-dimensional benchmark systems and demonstrated exceptional performance, achieving orders of magnitude speedup over both brute force Jacobian calculation and the Hutchinson estimator while maintaining unbiased sampling of the Boltzmann distribution. Notably, our method achieved unbiased sampling of Chignolin protein\cite{Honda:2004vj, Satoh:2006us} with explicit representation of all its atomic Cartesian coordinates. To the best of our knowledge, this marks the largest molecule ever Boltzmann sampled with such detailed representation using generative models. This advancement highlights the potential of our approach for larger and more complex molecular systems.

\section*{Flow Perturbation}

Flow-based generative models, such as NF and CNF, generate configurations of the target system by first sampling from a simple prior distribution and then transforming these samples through an invertible and differentiable map, denoted as \( \f: \z \rightarrow \x \). With appropriately trained \( \f \), these models can generate configurations of complex many-body systems such as proteins. However, the configurations they generate do not necessarily unbiasedly sample from the target Boltzmann distribution, due to reasons such as limited capacity of the flow model, insufficient training, biases in the training data, etc\cite{Noe:2019uu}.

\begin{figure}[h]
    \centering
    \includegraphics[width=\linewidth]{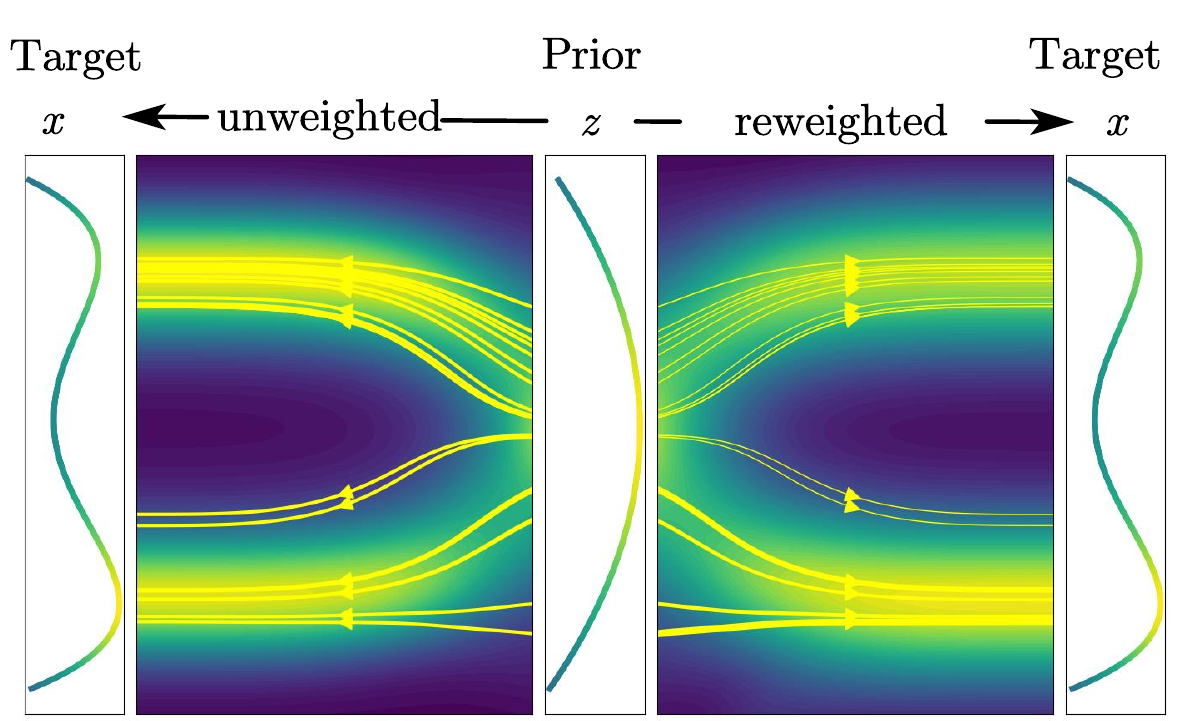}
    \caption{ Reweighting trajectories generated by a flow model. The prior distribution, shown in the middle, is Gaussian, while the target distribution features two peaks of different heights. The original trajectories, shown on the left, fail to sample the target distribution correctly, with roughly equal numbers of trajectories reaching each peak despite their differing heights. On the right, the trajectories are reweighted according to $e^{-W}$, resulting in trajectories that arrive at the lower probability regions being given less weight. This reweighting ensures unbiased sampling of the target distribution.}
    \label{fig:reweighting}
\end{figure}

To ensure unbiased sampling, one need to reweight the trajectories generated by the flow model according to the exponential of the negative generalized work~\cite{Noe:2019uu, SNF_Wu2020}, \( e^{-W} \), as illustrated in Figure \ref{fig:reweighting}. The generalized work \( W \) of a trajectory $\z \to \x$ is defined as:
\begin{equation}
W = u_X(\x) - u_Z(\z) - \Delta S \,,
\label{eq:work}
\end{equation}
Here \( u_X(\x) \) and \( u_Z(\z) \) represents the dimensionless potential energy of the Boltzmann distribution $\frac{1}{Z_X} \exp(- u_X(\x))$ and the prior distribution $q(\z) = \frac{1}{Z_Z} \exp(- u_Z(\z))$, respectively. The entropy term \( \Delta S \) is given by the Jacobian determinant of the flow,
\begin{equation}
\Delta S = \log \left| \det \left( \frac{\partial \f}{\partial \z} \right) \right|\,.
\end{equation}
 Calculating the entropy term is challenging and computationally intensive, especially for high-dimensional systems. This is because it requires \( D \) backpropagation passes through the flow, where \( D \) is the dimension of the system.

To address this computational challenge, we introduce the flow perturbation method. This method is based on the fact that the reweighting scheme applies not only to deterministic flows but also to stochastic processes. For stochastic trajectories, reweighting is performed using the exponential of the negative work, just as in the deterministic case. However, the key difference is that the entropy term for a stochastic trajectory is determined by the conditional probability ratio between this trajectory with its corresponding reverse trajectory\cite{SNF_Wu2020, doi:10.1073/pnas.1106094108}, which offers a potential way to avoid the computationally intensive Jacobian calculation.

\begin{figure}[h]
    \centering
    \includegraphics[width=\linewidth]{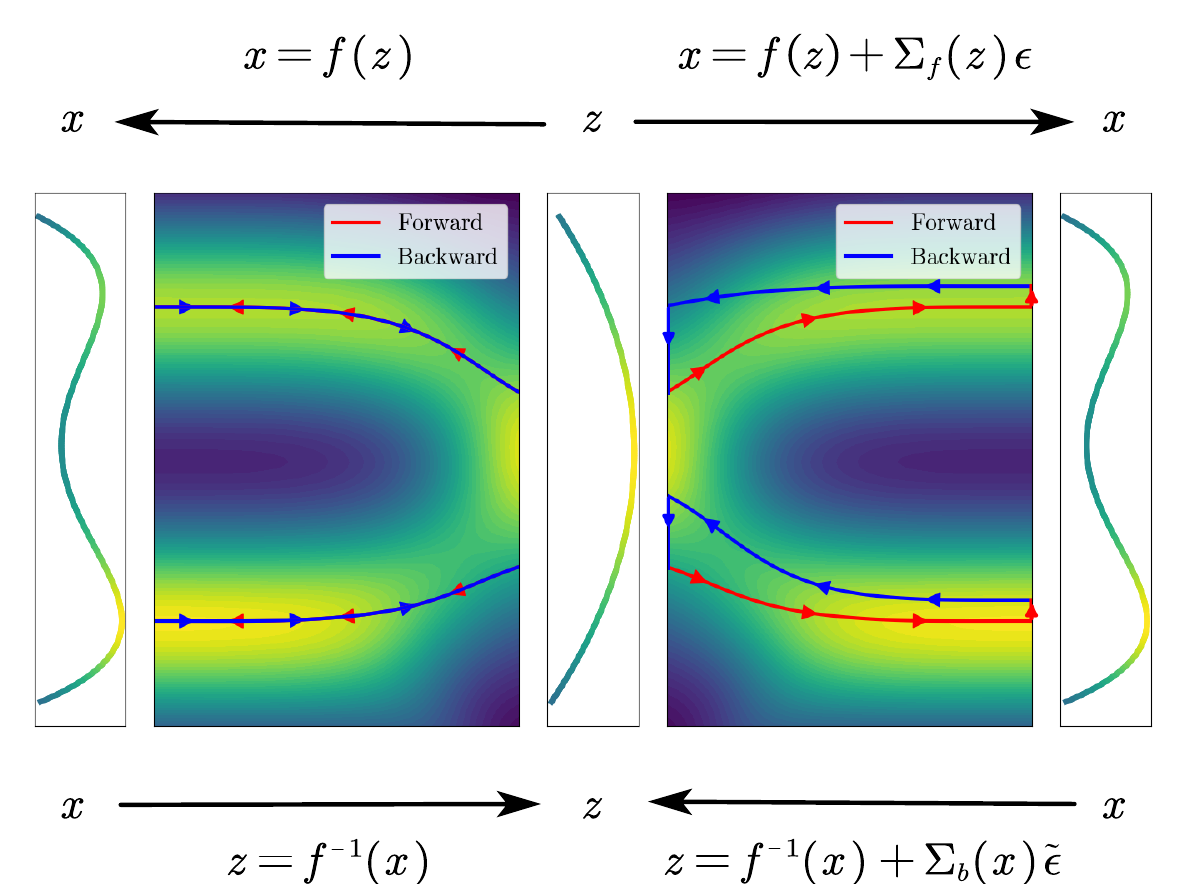}
    \caption{An illustration of the flow perturbation method. The left panel shows the forward and corresponding backward trajectories generated by the original flow model, while the right panel shows the trajectories generated by the perturbed flow model.}
    \label{fig:fp}
\end{figure}

To implement the flow perturbation method, we add the following stochastic perturbations to the flow and the inverse flow, which create a forward and backward stochastic process:
\begin{eqnarray}
\text{Forward:} & \x &= \f(\z) + \Sigma_f(\z) \epsilon \label{eq:fwd}\\
\text{Backward:} & \z &= \f^{-1}(\x) + \Sigma_b(\x) \tilde{\epsilon} \,.\label{eq:bwd}
\end{eqnarray}
Here \( \epsilon \) and \( \tilde{\epsilon} \) are random noises drawn from a standard Gaussian distribution, and \(\Sigma_f(\z) \) and \( \Sigma_b(\x) \) are matrices that scales the forward and backward noise, respectively. An illustration of the perturbed flow is shown in Figure \ref{fig:fp}.

With the forward and backward process defined in Eq.(\ref{eq:fwd}) and (\ref{eq:bwd}), the entropy of a forward trajectory $\z \to \x$ is given by 
\begin{equation}
\Delta S = \frac{\|\epsilon\|^2 - \|\tilde{\epsilon}\|^2}{2} + \log \left(\left| \frac{\det\Sigma_f(\z)} {\det\Sigma_b(\x)}\right| \right ) \,, \label{eq:deltaS}
\end{equation}
which is obtained by taking the negative logarithm of the conditional probability ratio between this forward path $\z \to \x$ with the corresponding backward path $\x \to \z$. A detailed derivation of this entropy term can be found in Supplementary Sec.~\ref{appendix:entropy}. 

The entropy term in Eq.(\ref{eq:deltaS}) no longer involves Jacobian calculation. However, this will not necessarily lead to an acceleration of the reweighting process. This is because  the entropy term now includes an additional component, \( \frac{\|\epsilon\|^2 - \|\tilde{\epsilon}\|^2}{2} \), which introduces extra fluctuations. Furthermore, the energy term of the generalized work \( W \), which is  still determined by difference in dimensionless potential energies, also experiences increased fluctuations due to the perturbations introduced by the forward noise, which affects the quality of generated configurations.

The increased fluctuations of $W$ will reduce the efficiency of the reweighting scheme. Specifically, its efficiency decreases exponentially with increasing $W$ fluctuations, meaning that exponentially more trajectories are needed to achieve the same accuracy in the expectation values of thermodynamic observables\cite{Pohorille:2010ue}. This additional computational cost could negate the benefit of avoiding Jacobian calculations.

To minimize the additional fluctuations in $W$ introduced by the noises, we impose  restrictions on the choice of scaling matrices $\Sigma_f(\z)$ and $\Sigma_b(\x)$. First, to reduce the additional fluctuation in the energy term, we set the forward scaling matrix $\Sigma_f(\z)$ to be a small constant:
\begin{equation}
\Sigma_f(\z) = \sigma_f \mathbf{I} \,.
\end{equation}
The constant $\sigma_f$ should be small enough to make sure the energy of configurations generated are minimally affected by the noise. However, $\sigma_f$ must not be too small that the forward perturbation becomes negligible and is disregarded as a rounding error in numerical calculations.

Second, to reduce the additional fluctuation in the entropy term, we parameterize the backward scaling matrix $\Sigma_b(\x)$ as a neural network model and train it according to the following loss function:
\begin{equation}
    \text{Loss} = \mathbb{E}_{\z \to \x} \left| \|\epsilon\|^2 - \|\tilde{\epsilon}\|^2 \right| \,.
\label{eq:loss}
\end{equation}
This loss represents  the absolute difference between $\|\epsilon\|^2$ and $\|\tilde{\epsilon}\|^2$,  averaged over all possible forward trajectories.   Minimizing this loss will directly reduce the fluctuation introduced by the first term of \(\Delta S\). 


 This loss function can in principle be minimized to exactly zero when $\sigma_f$ is infinitely small, given that the neural network
for $\Sigma_b(\x)$ has enough modeling capacity and the training is perfect (Supplementary Sec.~\ref{appendix:zero}). However, implementing such a neural network, which has $D^2$ output nodes, would require huge amount of computational resources.

For practical considerations, we approximate the $\Sigma_b(\x)$ as a scalar function:
\begin{equation}
\Sigma_b(\x) = \sigma_b(\x) \mathbf{I}  \,.
\end{equation}
Here $\sigma_b(\x)$ is modeled as a neural network that takes a $D$-dim vector as input and output a scalar, which will be much less expensive to train and use for reweighting. Although the loss function can no longer be minimized to zero with this approximation, it usually can be reduced sufficiently to ensure that the additional fluctuations minimally affect the reweighting efficiency.

A detailed description of the training algorithm for $\sigma_b(\x)$ can be found in Algorithm \ref{alg:sigb_training}.
\begin{algorithm}
    \caption{Training \(\sigma_b(\x)\)}\label{alg:sigb_training}
    \begin{algorithmic}[1]
        \STATE Initialize parameters \(\theta\) for $\sigma_b(\x;\theta)$
        \STATE Set learning rate $\eta$
        \REPEAT
            \STATE Sample \(\epsilon \sim \mathcal{N}(0, \mathbf{I})\), \(\z \sim q(\z)\)
            \STATE \(\x \gets \f(\z) + \sigma_f \epsilon\)        
            \STATE \(\tilde{\epsilon} \gets \frac{1}{\sigma_b(\x;\theta)}(\z - \f^{-1}(\x)) \)
            \STATE  \(\theta \gets \theta -\eta \nabla_\theta \left| \|\epsilon\|^2 - \|\tilde{\epsilon}\|^2 \right|\) 
        \UNTIL{converged}
    \end{algorithmic}
\end{algorithm}


\section*{Metropolis MC with Perturbed Flow}
In this study, we integrate the perturbed flow with Metropolis Monte Carlo (MC) method to ensure that trajectories are correctly reweighted by \( e^{-W} \). Metropolis MC is often combined with flow models for sampling Boltzmann distributions\cite{PhysRevD.100.034515, Sbailo:2021vr, Gabrie:2022wf, schonle:hal-04404948, Wang:2024vq}. Unlike the Importance Sampling (IS) method\cite{Tokdar:2010va}, which directly assigns weights to trajectories\cite{Noe:2019uu}, Metropolis MC uses an acceptance-rejection mechanism to accept new trajectories based on their importance weights.  Compared to IS, Metropolis MC is capable of performing incremental updates to the trajectories, which allows for more thorough exploration of the trajectory space, especially in regions that are poorly covered by the flow.

During each MC step, we perform partial updates\cite{PhysRevD.104.114507, schonle:hal-04404948}, instead of full updates, on the random variables of a trajectory, in order to maintain reasonably high acceptance rate. Specifically, we randomly select a subset of the dimensions of the latent variable \( \z \) and the forward noise variable \( \epsilon \) and resample these dimensions from the prior distribution and the standard Gaussian distribution, respectively.   It is important to note that partial updates of \( \z \) can only be achieved when the dimensions of the prior distribution are independent of each other, which is typically the case since prior distributions are usually simple distributions such as Gaussian or uniform.

With the partially updated variables, denoted as \( \z' \) and \( \epsilon' \), we obtain a new configuration \( \x' \) following Eq.(\ref{eq:fwd}). This new trial trajectory $\z' \to \x'$ is then accepted or rejected according to the Metropolis acceptance criterion, which is defined as follows:
\begin{equation}
\text{acc} = \min \left( 1, \frac{e^{-W(\z' \rightarrow \x')}}{e^{-W(\z \rightarrow \x)}} \right) \,.
\end{equation}
Here \( \z \rightarrow \x \) refers to the current trajectory, and $W(\z' \rightarrow \x')$ and $W(\z \rightarrow \x)$ refer to the generalized work of the trial and current trajectory, respectively. This acceptance criterion ensures detailed balance, guaranteeing that configurations generated by Metropolis MC unbiasedly sample the Boltzmann distribution (as shown in the Supplementary Sec.~\ref{appendix:DB}).

\section*{Results}

We assessed the performance of our flow perturbation (FP) method in sampling the Boltzmann distribution of two benchmark systems: the Gaussian Mixture Model (GMM) and the Chignolin protein. We conducted Metropolis MC sampling using the perturbed flow and compared its efficiency and accuracy to using the base flow model \( \f \) directly in MC (see Supplementary Sec.~\ref{appendix:MCflow}), in which case the Jacobian of the base flow is obtained using two different approaches: brute force evaluation and the Hutchinson estimator.

The brute force evaluation of Jacobian (BFJacob), while performs \( D \) backpropagation passes through the flow with Automatic Differentiation framework to derive the Jacobian matrix (Supplementary Sec.~\ref{appendix:Jacob}),  is accurate but computationally intensive. The Hutchinson estimator, on the other hand, approximates the trace of the Jacobian matrix by averaging the products of the Jacobian matrix with random vectors. Increasing the number of  random vectors used by the estimator  increases its accuracy  but also increases the computational cost. In fact, one needs to perform $N$ backpropgation passes if $N$ random vectors are used by the estimator (Supplementary Sec.~\ref{appendix:Jacob}). In this study, we tested the setup of using a single random vector (Hutch1) and using ten random vectors (Hutch10) to estimate the trace.

\subsection*{Benchmark Systems}

\subsubsection*{GMM}
The Gaussian Mixture Model (GMM) is a probabilistic model that represents a distribution as a combination of multiple Gaussian distributions. In this study,  a 1000-dimensional GMM is used as the benchmark system, which consisting of 10 Gaussian components. The Gaussian components are chosen to be well separated from each other.  This setup creates a complex distribution with well-separated modes, which provides a stringent test for the sampling methods.  Further details about the GMMs used in our tests can be found in Supplementary Sec.~\ref{appendix:GMM}.

\subsubsection*{Chignolin}
Chignolin\cite{Honda:2004vj, Satoh:2006us} is a small peptide consisting of 10 amino acid residues (GYDPETGTWG). It is one of the smallest peptides with a well-defined folded structure, specifically a beta hairpin, making it an ideal model for studying protein folding dynamics.

\begin{figure}[h]
    \centering
    \includegraphics[width=0.8\linewidth]{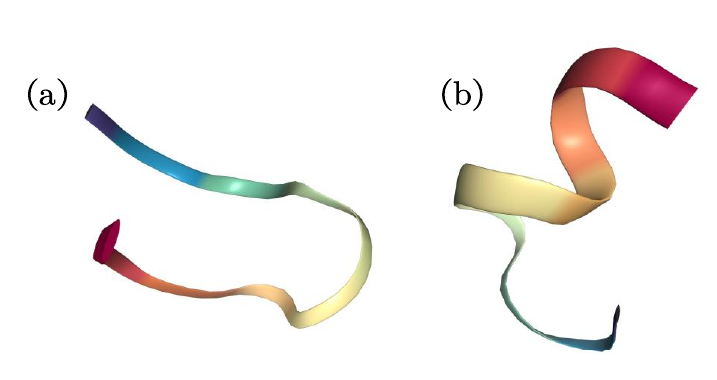}
    \caption{The representative configurations of Chignolin at 300K: (a)  beta-hairpin and (b) a configuration with partial alpha-helical segment.}
    \label{fig:struc}
\end{figure}

In this study, the Cartesian coordinates of all the atoms of Chignolin (a total of 175 atoms) are modeled directly. The energy of Chignolin is determined using the CHARMM22 force field\cite{MacKerell-Jr.:2000tn} with the implicit OBC2 solvent model\cite{Onufriev:2004vx}. We used simulation data from Frank Noé's research group\cite{chignolin_simulation_data}, who performed replica-exchange molecular dynamics study of Chignolin with this force field, as our training set for the flow model. Specifically, we utilized configurations at 300K for training. The representative configurations of Chignolin protein at this temperature are shown in Figure \ref{fig:struc}.
\subsection*{Implementation Details}
For both systems, the base flow models were implemented as Probability-Flow ODEs of the diffusion model\cite{song2021scorebased,NEURIPS2022_a98846e9} and trained using denoising score matching\cite{song2021scorebased, NEURIPS2020_4c5bcfec}. The score functions were modeled as multi-layer perceptrons (MLPs) with residual connections and sinusoidal time embedding. Details about the neural network architecture and the training settings are provided in Supplementary Sec.~\ref{appendix:flow_used}.

The forward scaling constant (\(\sigma_f\)) was set to 0.01 for the GMM and 0.001nm for Chignolin. Other choices of \(\sigma_f\)  are also tested, with the results summarized in Supplementary Fig.~\ref{fig:diffsig_gmm} and \ref{fig:diffsig_cgn}. As demonstrated in the figure, \(\sigma_f\) can vary over a wide range without impacting the efficiency and accuracy of the sampling.

The backward scaling function (\(\sigma_b(\x)\)) was also modeled as an MLP with residual connections. Details for this model are included in the Supplementary Sec.~\ref{appendix:bnv_used}.

To maintain reasonable acceptance rates in MC, we performed partial updates for the random variable $\z$ and $\epsilon$ as described earlier.

For the GMM, we updated 5 coordinates of $\z$ and $\epsilon$ per trial move, while for the Chignolin protein, we updated 1 coordinate per trial move. We also explored different numbers of updated coordinates, with detailed results provided in Supplementary Fig.~\ref{fig:diffK_gmm} and \ref{fig:diffK_cgn}. For the BFJacob, Hutch1, and Hutch10 methods, the same number of coordinates of \(\z\) were updated as in the FP method to ensure a fair comparison. 

\subsection*{Computational Efficiency}
The total computational cost of a MC simulation is determined by the computational cost of a single MC step and the number of steps taken for the MC to converge. Therefore we evaluate the efficiency of the FP, BFJacob, Hutch1, and Hutch10 methods in these two aspects. For the FP method, we also benchmarked the time taken to train \(\sigma_b(\x)\), as this training time is unique to FP and should be included for a fair comparison.

\begin{figure*}[!htb]
	\centering
	\includegraphics[width=1.0\textwidth]{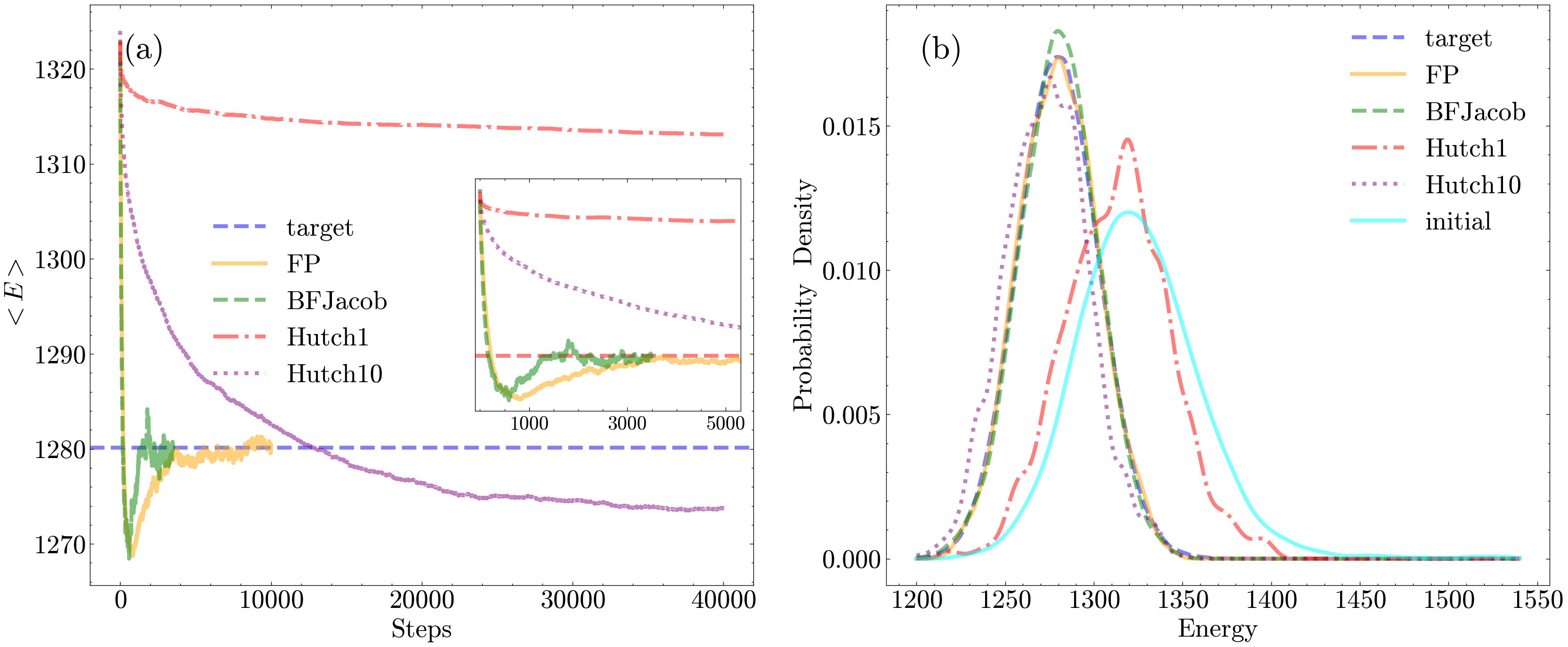}
	\caption{Sample the Boltzmann distribution of the 1000-dimensional GMM using different methods.
    (a) Average energy \(\left<E\right>\) as a function of MC steps obtained with different methods. The target \(\left<E\right>\) is marked by a blue dashed line.
    (b) Energy distributions obtained using different methods, compared to the target distribution. The initial distribution, representing the energy distribution of samples generated directly by the base flow model, is shown for reference.
    }
    \label{fig:gmm}
\end{figure*}

The computational time required to perform a single MC step with the FP, BFJacob, Hutch1, and Hutch10 methods are presented in Table \ref{table:gmm} and \ref{table:cgn}. For both the 1000-dimensional GMM and Chignolin system, the BFJacob
method demands hundreds of times more computational time per MC step
compared to the FP method.  Hutch1 consumes a similar amount of computational time per MC step as the FP method, while Hutch10 requires several times more computational time per step.

\begin{figure*}[!htb]
	\centering
	\includegraphics[width=1.0\textwidth]{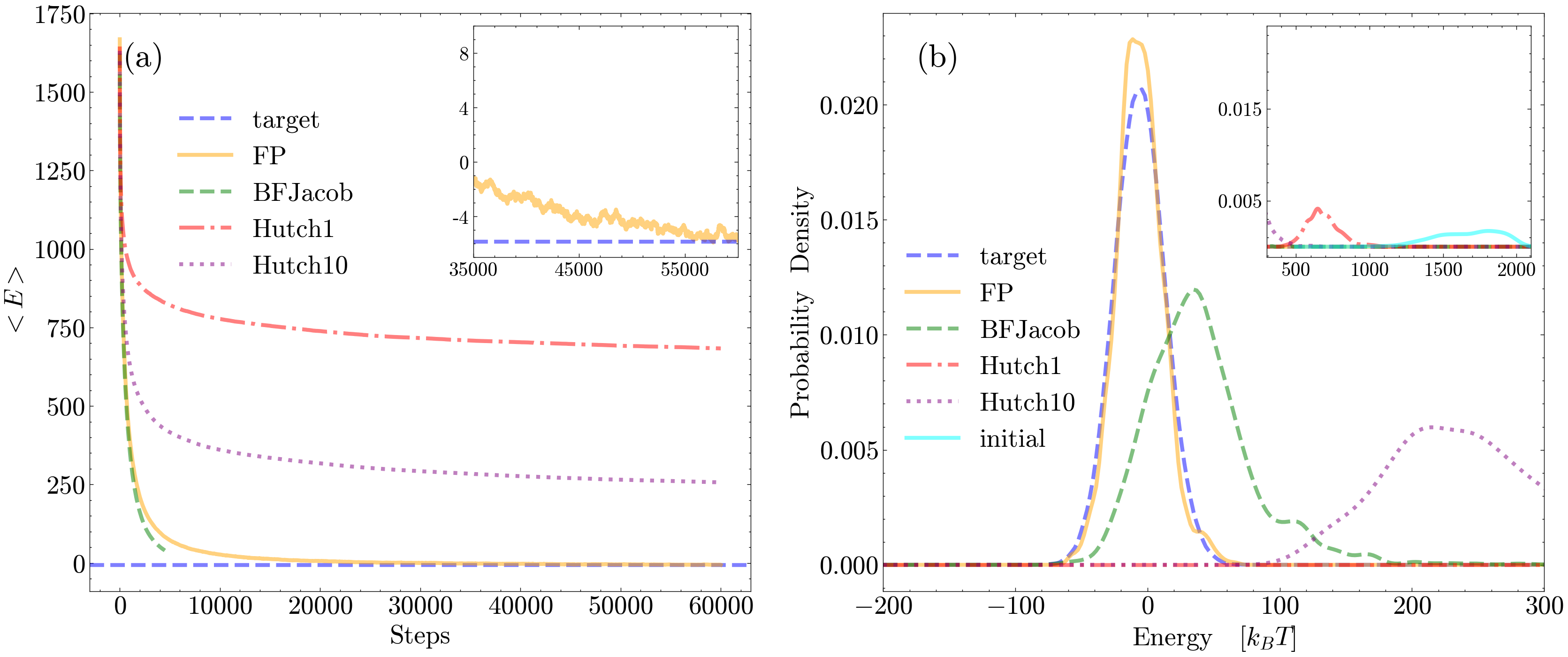}
	\caption{Sampling the Boltzmann distribution of the Chignolin using different methods.
    (a) Average energy \(\left<E\right>\) as a function of MC steps obtained with different methods. The target \(\left<E\right>\) is marked by a blue dashed line.
    (b) Energy distributions obtained using different methods, compared to the target distribution. The initial distribution, representing the energy distribution of samples generated directly by the base flow model, is shown for reference.
    }
    \label{fig:cgn}
\end{figure*}

The average energy of samples (\(\left<E\right>\)) is used as the indicator to determine the convergence of the MC simulation. Figure \ref{fig:gmm}(a) and \ref{fig:cgn}(a) illustrates \(\left<E\right>\) as a function of MC steps obtained with different methods. As one can see, \(\left<E\right>\) obtained with Hutch1 and Hutch10 decays very slowly. For the 1000-dim GMM, \(\left<E\right>\) continues to decrease after 40,000 MC steps with the Hutch methods.  Similarly, for the Chignolin protein, \(\left<E\right>\) is still decreasing even after 60,000 MC steps with these methods.  This slow convergence aligns with previous findings\cite{pmlr-v119-kohler20a, erives2024verletflowsexactlikelihoodintegrators} that the Hutchinson estimator requires much more samples for the reweighting scheme to converge due to the extra fluctuations in generalized work introduced by the stochastic trace estimator. One could, in principle, use more random vectors in the estimator to reduce these fluctuations, but that would significantly increase the computational cost of each MC step.

\begin{table}[H]
    \centering
    \caption{Computational cost of different methods for GMM. Hutch1 and Hutch10 did not converge after 40,000 MC steps. For the FP method, the training time for $\sigma_b(x)$ (0.07 hours) is included in the total time cost. All benchmarks were performed on a RTX 3090 GPU.}
    \begin{tabular}{c c c c}
        \hline
        \textbf{Method} & \textbf{1 Step} & \textbf{Steps} & \textbf{Time}  \\
        \hline
        \textbf{FP} & 2.7s & $\approx$3.5k & \textbf{0.07h}+2.6h \\
        \textbf{BFJacob} & 486.6s & $\approx$2k & 270h  \\
        \textbf{Hutch1} & 2.6s & $>$40k & $>$29h  \\
        \textbf{Hutch10} & 13.6s & $>$40k & $>$152h \\
        \hline
    \end{tabular}
    \label{table:gmm}
\end{table}

In contrast, the average energy \(\left<E\right>\) obtained with the FP method converges much faster. For the 1000-dim GMM, \(\left<E\right>\) reaches the target value after 3500 MC steps, while for the Chignolin protein, it converges to the target value after 55000 steps. This much faster convergence compared to Hutchinson methods suggests that, although the FP method also introduces extra fluctuations into the generalized work, these fluctuations are effectively reduced by choosing and training $\sigma_f$ and $\sigma_b(x)$ following the previously described procedure. 

\begin{table}[H]
    \centering
    \caption{Computational cost of different methods for Chignolin. BFJacob did not converge after 4500 MC steps. Hutch1 and Hutch10 did not converge after 60,000 MC steps. For the FP method, the training time for $\sigma_b(x)$ (0.5 hours) is included in the total time cost. All benchmarks were performed on a RTX 3090 GPU.}
    \begin{tabular}{c c c c}
        \hline
        \textbf{Method} & \textbf{1 Step} & \textbf{Steps} & \textbf{Time}  \\
        \hline
        \textbf{FP} & 2.4s & $\approx$55k & \textbf{0.5h}+36.7h \\
        \textbf{BFJacob} & 245.3s & $>$4.5k & $>$ 307h  \\
        \textbf{Hutch1} & 2.4s & $>$60k & $>$ 40h  \\
        \textbf{Hutch10} & 8.2s & $>$60k & $>$ 137h \\
        \hline
    \end{tabular}
    \label{table:cgn}
\end{table}

 The average energy \(\left<E\right>\) obtained with BFJacob decays slightly faster compared to the FP, as shown in Figure \ref{fig:gmm}(a) and \ref{fig:cgn}(a). For the 1000-dim GMM, \(\left<E\right>\) obtained with BFJacob reaches the target value after 2000 MC steps. For the Chignolin protein,  due to computational resource constraints, BFJacob was only executed for 4500 MC steps, and \(\left<E\right>\) obtained decays slightly faster than with FP. This is expected because, while the extra fluctuations in work introduced by FP have been effectively reduced, they are not completely eliminated, causing the FP method to require more MC steps to converge. 

Overall, the FP method  reduces the total computational cost by orders of magnitude compared to the BFJacob and Hutch methods, even when accounting for the additional time required to train $\sigma_b(x)$. For instance, in the GMM system, BFJacob consumes 100 times more computational time than FP, while Hutch1 and Hutch10 consume 10 and 50 times more, respectively, as shown in Table \ref{table:gmm}. It is important to note that these numbers for Hutch1 and Hutch10 are conservative estimates, as they have not yet converged. As for the Chignolin protein, BFJacob and Hutch methods are still far from converging, making it impossible to precisely estimate their computational costs compared to FP. However, it is expected that their computational costs will be similarly high as observed in the GMM case.

\subsection*{Sampling Accuracy}
The energy distribution of samples for the 1000-dimensional GMM obtained using different methods are illustrated in Figure \ref{fig:gmm}(b). Both the FP and BFJacob methods achieve distributions that closely match the target. In contrast, the Hutch1 method exhibits significant deviations from the target distribution, while the Hutch10 method shows a slight leftward shift, resulting in a lower average energy value than the target. It is important to note that the Hutchinson methods have not yet fully converged, and their energy distributions may further change with additional MC steps.

\begin{figure}[H]
	\centering
	\includegraphics[width=0.5\textwidth]{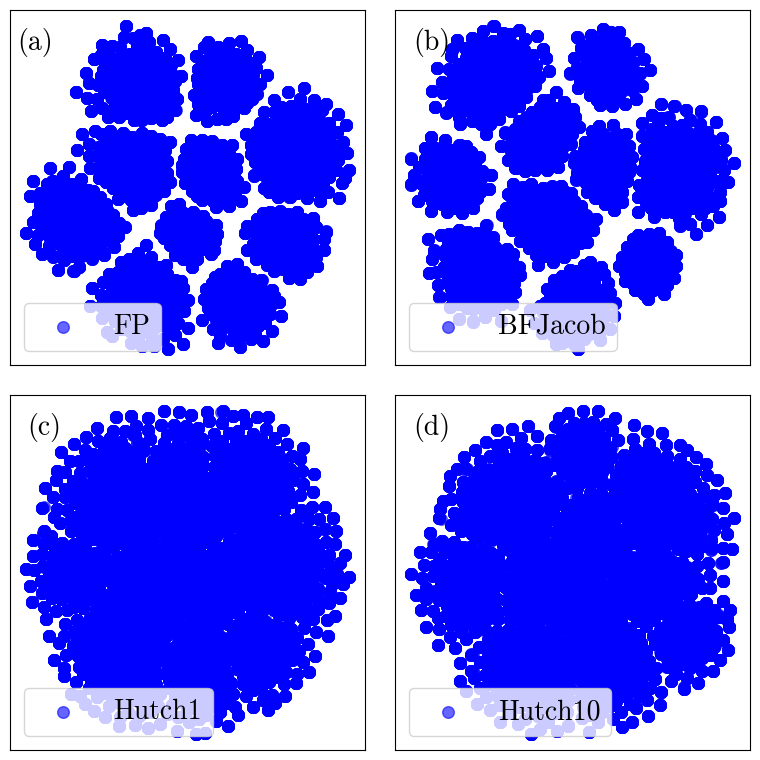}
	\caption{T-SNE visualization of the samples for 1000-dimensional GMM obtained with different methods. (a) FP, (b) BFJacob, (c)Hutch1, (d) Hutch10.
    }
    \label{fig:tsne}
\end{figure}

The t-SNE plots of the samples generated by different methods for the 1000-dimensional GMM are shown in Figure \ref{fig:tsne}. These plots provide a visual representation of how well each method separates samples into distinct modes. Both the FP and BFJacob methods successfully separate samples into 10 distinct modes as desired, while Hutch1 and Hutch10  fail to achieve mode separation.

\begin{figure}[h]
    \centering
    \includegraphics[width=0.8\linewidth]{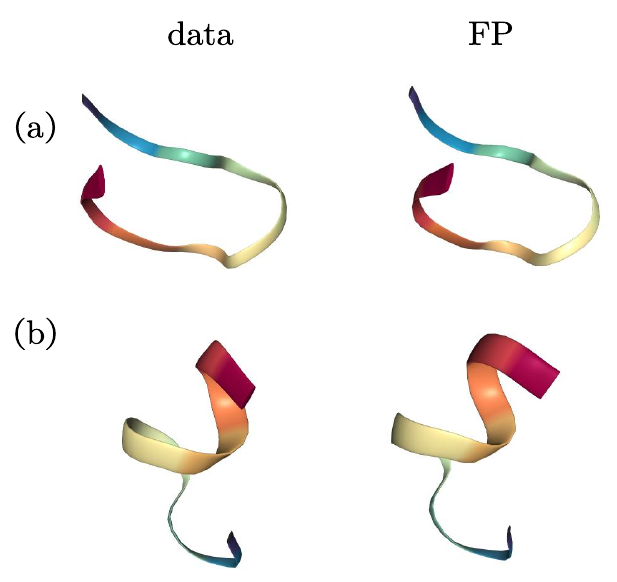}
    \caption{Representative configurations of Chignolin obtained from training data and Flow Perturbation method: (a) beta-hairpin, and (b) a configuration with partial alpha-helical segment.}
    \label{fig:struc_comp}
\end{figure}

The energy distribution of samples for Chignolin obtained by different methods is shown in Figure  \ref{fig:cgn}(b).  The FP method is able to  accurately reproduce the target distribution. Also, both representative structure of Chignolin are recovered by FP,  as shown in Figure \ref{fig:struc_comp}. The energy distributions obtained using Hutch1 and Hutch10 remain far from the target distribution.  The energy distribution achieved with BFJacob does not align with the target distribution either, although one needs to keep in mind that only 4500 MC steps have been run with BFJacob due to the extended duration of each step.

\section*{Discussion}
In this study, we introduced the flow perturbation method to accelerate sampling the Boltzmann distribution of high-dimensional systems. Our approach injects stochastic perturbations into flow-based generative models, eliminating the need to calculate the Jacobian of the flow. By adopting a small forward noise scaling constant and training the backward noise scaling function to minimize entropy fluctuations, the reweighting efficiency can be significantly improved.

When tested on the 1000-dimensional GMM and the Chignolin protein, our FP method demonstrated orders of magnitude acceleration compared to both brute force Jacobian evaluation and the Hutchinson estimator. Specifically, the FP method is hundreds of times faster per MC step than BFJacob. Moreover, the FP method requires only slightly more MC steps to converge compared to BFJacob, which is much less than what is needed by the Hutchinson method. This makes the FP method a highly efficient approach for sampling the Boltzmann distribution of complex many-body systems.

A limitation of the flow perturbation approach is that, while it significantly improves efficiency, it does not inherently increase accuracy compared to  the brute force Jacobian evaluation. This can be problematic if the base flow model experiences mode collapsing. In such cases, neither performing MC using the base flow with brute force Jacobian evaluation nor performing MC using the perturbed flow is likely to recover the missing modes.

To mitigate this issue, one might want to explore the possibility of adding different types of perturbations to the flow. For instance, employing noises with longer tail than the Gaussian noise, such as Cauchy noise, might facilitate the exploration of the target space and potentially aid in the recovery of these elusive modes.

\section*{Code Availability}
All codes used in this work are available on Github:\\
\url{https://github.com/XinPeng76/Flow_Perturbation}

\section*{Acknowledgement}
This research was supported by National Natural Science Foundation of China under grant number 12204059.

\section*{Author Contribution}
A.G. designed the study. X.P. and A.G. wrote the code, performed calculations, analyzed data, and wrote the manuscript.

\section*{Competing Interests}
The authors declare no competing interests.


\printbibliography

\onecolumn
\appendix
\setcounter{equation}{0}
\setcounter{figure}{0}
\addcontentsline{toc}{chapter}{Supplementary Material}
\largerAppendixTitle{Supplementary Material}
\section{Entropy Term of the Perturbed Flow \label{appendix:entropy}}
In this section, we derive the entropy term in Eq.(\ref{eq:deltaS}) of the main text.

For a stochastic process, the entropy of an arbitrary trajectory \( \z \rightarrow \x \) is defined as the negative logarithm of the ratio between the conditional probabilities of the forward and backward paths\cite{SNF_Wu2020, doi:10.1073/pnas.1106094108}.

The conditional forward path probability $P_f(\x | \z)$ for the forward process in Eq.(\ref{eq:fwd}) is
\begin{eqnarray}
    P_f({x} \mid {z}) & = & \frac{1}{(2\pi)^{D/2} |\det \Sigma_f({z})|} \exp\left( -\frac{1}{2} ({x} - \f({z}))^T (\Sigma_f({z}) \Sigma_f({z})^T)^{-1} ({x} - \f({z})) \right) \nonumber \\
    & = & \frac{1}{(2\pi)^{D/2} |\det \Sigma_f({z})|} \exp\left( -\frac{\norm{\epsilon}^2}{2} \right)
\end{eqnarray}

The conditional backward path probability $P_b(\z | \x)$ for the backward process in Eq.(\ref{eq:fwd}) is
\begin{eqnarray}
    P_b({z} \mid {x}) & = & \frac{1}{(2\pi)^{D/2} |\det \Sigma_b({x})|} \exp\left( -\frac{1}{2} ({z} - \f^{-1}({x}))^T (\Sigma_b({x}) \Sigma_b({x})^T)^{-1} ({z} - \f^{-1}({x})) \right) \nonumber \\
    & = & \frac{1}{(2\pi)^{D/2} |\det \Sigma_b({x})|} \exp\left( -\frac{\norm{\tilde{\epsilon}}^2}{2} \right)
\end{eqnarray}

The entropy of an arbitrary trajectory \( \z\rightarrow \x \) is defined as the  negative logarithm of the ratio between the conditional probabilities of the forward and backward paths, which gives
\begin{eqnarray}
    \Delta S & = & -\log \left( \frac{P_f(\x|\z)}{P_b(\z|\x)} \right)\nonumber\\
             & = &  \frac{\|\epsilon\|^2 - \|\tilde{\epsilon}\|^2}{2} + \log \left(\left| \frac{\det\Sigma_f(\z)}{\det\Sigma_b(\x)} \right| \right ) \,.
\end{eqnarray}

\section{Detailed Balance of Metropolis MC with Perturbed flow \label{appendix:DB}}

In this section, we demonstrate that the Metropolis MC dynamics utilizing the perturbed flow defined in the main text will converge to the target distribution \(\exp(-u_X(\x)) P_b(\z | \x)\). If the MC converges to this distribution, the \(\x\) samples obtained will unbiasedly represent the Boltzmann distribution.

To prove that Metropolis MC will converge to this target distribution, we need to show that the detailed balance condition is satisfied. The proof is as follows.

Let \(\Gamma\) denote an arbitrary trajectory \(\z \rightarrow \x\). Let \(P_{prop}(\Gamma)\) represent the probability of generating a trajectory \(\Gamma\) with our perturbed flow, which is \(P_{prop}(\Gamma) = \exp(-u_Z(\z)) P_f(\x | \z)\).  We will use \(P_{target}(\Gamma)\) to denote the target distribution \(\exp(-u_X(\x)) P_b(\z | \x)\) we mentioned earlier.

The acceptance rule for the Metropolis MC defined in the main text is \(acc(\Gamma \rightarrow \Gamma') = \min\left(1, \frac{e^{-W(\Gamma')}}{e^{-W(\Gamma)}}\right)\).

The transition probability from \(\Gamma\) to \(\Gamma'\) is therefore \(P(\Gamma \rightarrow \Gamma') = P_{prop}(\Gamma') \, acc(\Gamma \rightarrow \Gamma')\), which is the probability of first sampling \(\Gamma'\) and subsequently accepting it.

The detailed balance condition we want to prove is \(P_{target}(\Gamma) P(\Gamma \rightarrow \Gamma') = P_{target}(\Gamma') P(\Gamma' \rightarrow \Gamma)\). To prove it is satisfied, we need to consider separately the cases where \(\frac{e^{-W(\Gamma')}}{e^{-W(\Gamma)}} < 1\) and \(\frac{e^{-W(\Gamma')}}{e^{-W(\Gamma)}} \geq 1\).

When \(\frac{e^{-W(\Gamma')}}{e^{-W(\Gamma)}} < 1\), we have \(acc(\Gamma \rightarrow \Gamma') = \frac{e^{-W(\Gamma')}}{e^{-W(\Gamma)}}\) and \(acc(\Gamma' \rightarrow \Gamma) = 1\).

Utilizing the fact that

\begin{eqnarray}
\frac{e^{-W(\Gamma')}}{e^{-W(\Gamma)}} = \frac{P_{target}(\Gamma')}{P_{prop}(\Gamma')} \cdot \frac{P_{prop}(\Gamma)}{P_{target}(\Gamma)},
\end{eqnarray}

we obtain:

\begin{eqnarray}
P_{target}(\Gamma) P(\Gamma \rightarrow \Gamma')  & = & P_{target}(\Gamma) P_{prop}(\Gamma') \frac{e^{-W(\Gamma')}}{e^{-W(\Gamma)}}\nonumber\\
 & = & P_{target}(\Gamma) P_{prop}(\Gamma') \frac{P_{target}(\Gamma')}{P_{prop}(\Gamma')} \cdot \frac{P_{prop}(\Gamma)}{P_{target}(\Gamma)} \nonumber\\
 & = & P_{target}(\Gamma') P_{prop}(\Gamma)\nonumber\\
& = & P_{target}(\Gamma') P(\Gamma' \rightarrow \Gamma),
\end{eqnarray}

which proves the detailed balance in this case.

In the case where \(\frac{e^{-W(\Gamma')}}{e^{-W(\Gamma)}} \geq 1\), we have \(acc(\Gamma' \rightarrow \Gamma) = \frac{e^{-W(\Gamma)}}{e^{-W(\Gamma')}}\) and \(acc(\Gamma \rightarrow \Gamma') = 1\).

Starting from \(P_{target}(\Gamma') P(\Gamma' \rightarrow \Gamma)\), we have:

\begin{eqnarray}
P_{target}(\Gamma') P(\Gamma' \rightarrow \Gamma)  & = & P_{target}(\Gamma') P_{prop}(\Gamma)\frac{e^{-W(\Gamma)}}{e^{-W(\Gamma')}}\nonumber\\
 & = & P_{target}(\Gamma') P_{prop}(\Gamma) \frac{P_{target}(\Gamma)}{P_{prop}(\Gamma)} \cdot \frac{P_{prop}(\Gamma')}{P_{target}(\Gamma')}\nonumber\\
 & = & P_{target}(\Gamma) P_{prop}(\Gamma')\nonumber\\
   & = & P_{target}(\Gamma) P(\Gamma \rightarrow \Gamma'),
\end{eqnarray}

which proves the detailed balance in this case.

With both cases proved, we have now completed the proof of the detailed balance.

\section{Exactly Minimizing The Loss Function \label{appendix:zero}}
To prove that the loss function in Eq.(\ref{eq:loss}) can be minimized to zero under the condition that $\sigma_f$ is infinitely small, we need to identify a specific $\Sigma_b(\x)$ such that the loss function evaluates to zero when this matrix is used.

A suitable matrix that satisfy this condition is:
\begin{equation}
\Sigma_b(\x) = \sigma_f \frac{\partial \f^{-1}}{\partial \x} \,,
\label{eq:SI_sigmab}
\end{equation}
where $\frac{\partial \f^{-1}}{\partial \x}$ is the Jacobian matrix of the inverse flow.

To prove this matrix minimize the loss function to zero, one need to use the fact that 
\begin{equation}
\tilde{\epsilon} = \Sigma_b(\x)^{-1}(\z - \f^{-1}(\x) ) .
\label{eq:SI_noise}
\end{equation}
This is the backward noise needed to make sure the trajectory goes exactly back to $\z$.

In the limit that $\sigma_f$ is infinitely small, we have
\begin{eqnarray}
\f^{-1}(\x) & = & \f^{-1}(\f(\z)+\sigma_f \epsilon) \nonumber \\
          & = & \z + \sigma_f \frac{\partial \f^{-1}}{\partial \x} \epsilon\,.
\label{eq:SI_taylor}
\end{eqnarray}

Substituting Eq.(\ref{eq:SI_sigmab}) and Eq.(\ref{eq:SI_taylor}) into Eq.(\ref{eq:SI_noise}), we obtain
\begin{equation}
    \tilde{\epsilon} = \epsilon \,,
\end{equation}
which indicate that with the specific choice of $\Sigma_b(\x)$ in Eq.(\ref{eq:SI_sigmab}), the backward noise $\tilde{\epsilon}$ and forward noise $\epsilon$ will be exactly the same as each other. In this case, the loss function is exactly zero.

\section{Metropolis MC with deterministic flow \label{appendix:MCflow}}
The samples generated by the flow-based generative model follow a distribution that is generally different from the target Boltzmann distribution \( e^{-u_X(\x)} \), where \( u_X(\x) \) represents the dimensionless potential energy. To achieve unbiased Boltzmann sampling, one can combine it with Metropolis MC, as detailed below. 

To  leverages flow-based models in MC, one uses the flow model to propose trial configurations and employs the Metropolis-Hastings algorithm to accept or reject these configurations, ensuring unbiased sampling from the Boltzmann distribution. At each MCMC step, a new configuration \( \x' \) is proposed by sampling \( \z' \) from the prior distribution and transforming it using the flow-based model \( \f \), such that \( \x' = \f(\z') \).

The proposed configuration \( \x' \) is accepted with a probability given by the Metropolis-Hastings acceptance rule:

\begin{equation}
\text{acc} = \min \left( 1, \frac{e^{-W(\z' \rightarrow \x')}}{e^{-W(\z \rightarrow \x)}} \right) \,,
\end{equation}
where
\begin{equation}
W(\z \rightarrow \x) = u_X(\x) - u_Z(\z) - \Delta S \,,
\label{eq:work}
\end{equation}
is the work performed along the original trajectory \( \z \rightarrow \x \). Here  $\Delta S = \log \left| \det \left( \frac{\partial \f}{\partial \z} \right) \right|$ corresponds to the entropy change along with the trajectory.

This acceptance rule satisfies the detailed balance condition\cite{Gabrie:2022wf}, which ensures that the MC dynamics correctly sample the target Boltzmann distribution upon convergence.

\section{Obtaining the Jacobian of CNF\label{appendix:Jacob}}
Continuous Normalizing Flows (CNFs) extend the concept of traditional NFs by defining the flow $\f: \z \to \x$ using an Ordinary Differential Equation (ODE):

\begin{equation} \frac{d\x}{dt} = {v}(\x(t), t), \quad \x(0) = \z \end{equation}

In this formulation, the mapping $\f$ is determined by a vector field \( {v}(\x(t), t) \), usually refered as velocity field. With this ODE-based approach, the log determinant of the Jacobian of the flow can be expressed as an integral of the divergence of the vector field:

\begin{equation} \log \left| \det \frac{\partial \f}{\partial \z} \right| = \int_0^1 dt \, \nabla \cdot {v}(\x(t), t) \end{equation}

The divergence of the velocity field \( {v} \) can be obtained as the trace of the Jacobian of \( {v} \).

In this section, we show two different ways of obtaining this Jacobian trace. The first way is to directly construct the Jacobian matrix using Automatic Differentiation(AD) framework like PyTorch, and then calculate the trace for this matrix. We denote this way as brute force calculation. The second way is to use the Hutchinson estimator to estimate this Jacobian trace.

\subsection{Constructing the Jacobian Matrix Using Automatic Differentiation }
Given an invertible and differentiable map \( v: {x} \rightarrow {y} \), where both \({x}\) and \({y}\) are \(D\)-dimensional vectors, we can construct the Jacobian matrix \(\frac{\partial v}{\partial {x}}\) using an Automatic Differentiation (AD) framework such as PyTorch or TensorFlow.

To achieve this, we need to compute the gradient of each component \( y_i \) of \({y}\) with respect to the input vector \({x}\). This involves several steps:

First, for each \(i\) (where \(i = 1, \ldots, D\)), we define a unit vector \({e}_i\) of dimension \(D\), with the \(i\)-th component equal to 1 and all other components equal to 0. This unit vector is used to isolate the \(i\)-th component of \({y}\). By computing the dot product of \({y}\) with \({e}_i\), we obtain:
\begin{equation}
y_i = {e}_i^T {y} = v_i({x}).
\end{equation}

Next, we perform backpropagation using an AD framework to compute the gradient of \( y_i \) with respect to \({x}\). This gradient gives the \(i\)-th row of the Jacobian matrix:
\begin{equation}
\nabla_{{x}} y_i = \left[ \frac{\partial y_i}{\partial x_1}, \frac{\partial y_i}{\partial x_2}, \ldots, \frac{\partial y_i}{\partial x_D} \right].
\end{equation}
Since \({y}\) is a \(D\)-dimensional vector, this process must be repeated \(D\) times, once for each component of \({y}\), thereby requiring \(D\) backpropagation steps.

Finally, we collect the gradients \(\nabla_{{x}} y_i\) for all \(i\) to form the Jacobian matrix:
\begin{equation}
\frac{\partial v}{\partial {x}} = \begin{bmatrix}
\frac{\partial y_1}{\partial x_1} & \frac{\partial y_1}{\partial x_2} & \cdots & \frac{\partial y_1}{\partial x_D} \\
\frac{\partial y_2}{\partial x_1} & \frac{\partial y_2}{\partial x_2} & \cdots & \frac{\partial y_2}{\partial x_D} \\
\vdots & \vdots & \ddots & \vdots \\
\frac{\partial y_D}{\partial x_1} & \frac{\partial y_D}{\partial x_2} & \cdots & \frac{\partial y_D}{\partial x_D}
\end{bmatrix}.
\end{equation}

In PyTorch 2.0, this process has been wrapped in function `jacrev`, which we utilized in this work to efficiently compute the Jacobian matrix. 

\subsection{Estimating the Jacobian Trace Using the Hutchinson Estimator}
The Hutchinson estimator is a stochastic method that provides an efficient way to estimate the trace of a matrix. It approximates the trace of an arbitrary matrix \( A \) by taking the expectation of the quadratic form \( {u}^T A {u} \), where \( {u} \) is a random vector drawn from a standard normal distribution. For the Jacobian matrix \( J = \frac{\partial v}{\partial \x} \), the trace can be estimated as follows.

First, we select a random vector \( {u} \) of the same dimensionality as the input space, with elements drawn from a standard normal distribution.

Next, we  compute the gradient of the scalar quantity \({u}^T v(\x)\) with respect to \(\x\) using AD framework, which gives us:
\begin{equation}
g = \nabla_{\x} ({u}^T v(\x)) = \frac{\partial ({u}^T v(\x))}{\partial \x}.
\end{equation}

Next, we compute the inner product \( {u}^T g \), which gives the required quadratic form \( {u}^T J {u} \).

This process needs to be repeated for multiple samples of \( {u} \). The trace of \( J \) can then be estimated by averaging the results:
\begin{equation}
\text{Tr}(J) \approx \frac{1}{N} \sum_{i=1}^N {u}_i^T \frac{\partial v}{\partial \x} {u}_i.
\end{equation}

As shown, this requires performing \(N\) backpropagation passes, where \(N\) is the number of random vectors used in this averaging process.

\section{GMM \label{appendix:GMM}}
The Gaussian Mixture Model (GMM) is a probabilistic model that represents a distribution as a mixture of multiple Gaussian distributions. In an n-dimensional GMM with \(k\) Gaussian components, the overall probability density \(p(\x)\) of the system can be expressed as:
\begin{equation}
p(\x) = \sum_{j=1}^{k} \pi_j \mathcal{N}(\x \mid \mu_j, \Sigma_j) \,,\label{eq:prob_GMM}
\end{equation}
where \(\pi_j\) is the mixing coefficient for the \(j\)-th Gaussian component, and \(\mathcal{N}(\x \mid \mu_j, \Sigma_j)\) represents the \(j\)-th multivariate Gaussian distribution.

For this study, we set all mixing coefficients \(\pi_j\) to be equal. Additionally, the covariance matrices \(\Sigma_j\) are diagonal matrices. The mean vectors \(\mu_j\) and the diagonal elements of \(\Sigma_j\) are randomly generated as follows:

- \(\mu_j\): sampled from \(\mathcal{N}(\mathbf{0}, \mathbf{I})\), where \(\mathbf{I}\) is the \(n\)-dimensional identity matrix.

- \(\Sigma_j\): diagonal matrix with diagonal elements sampled from \( 0.4 + |\mathcal{N}(0.1, 0.5)|\). 

- \(\pi_j\): set to \(1\) for all \(j\).

This probability distribution in Eq.(\ref{eq:prob_GMM}) can be rewritten in the form of a Boltzmann distribution:
\begin{equation}
p(\x) = \exp(-E(\x)) \,,
\end{equation}
where \(E(\x)\) is the energy of configuration \(\x\).

\section{Flow model for GMM and Chignolin \label{appendix:flow_used}}
The diffusion based generative model can be formulated as the probability flow ODE\cite{song2021scorebased,NEURIPS2022_a98846e9}:
\begin{equation}
d\x = \left( \frac{\dot{s}(t)}{s(t)} \x - s(t)^2 \dot{\sigma}(t) \sigma(t) \nabla_{\x} \log p\left( \frac{\x}{s(t)}; \sigma(t) \right) \right) dt \,.
\end{equation}
In this work, we will use this ODE as the base flow model for the flow perturbation approach. For the GMM and Chignolin, the scoring function   \(\nabla_{\x} \log p\left( \dfrac{\x}{s(t)}; \sigma(t) \right)\) and the scaling function $s(t)$ and $\sigma(t)$ are chosen differently, which we will discuss below.

\subsection{Flow model for GMM}
For the ODE used for GMM, we follow the design used by Karras et al\cite{NEURIPS2022_a98846e9}. 
Firstly, we choose $s(t)$ to be 1. Secondly, the $\sigma(t)$ is chosen to be $t$.
The scoring function  is parameterized as follows:
\begin{equation}
\nabla_{\x} \log p\left( \x; \sigma \right) =\left(D(\x;\sigma)-\x\right)/\sigma^2 \,,
\end{equation}
where  $D(\x;\sigma)$ is defined as
\begin{equation}
	D(\x;\sigma) = c_{\text{skip}}(\sigma)\x + c_{\text{out}}(\sigma)F_{\theta} \left( c_{\text{in}}(\sigma)\x;c_{\text{noise}}(\sigma) \right) \,.
\end{equation}
Here $F_{\theta} \left( c_{\text{in}}(\sigma)\x;c_{\text{noise}}(\sigma) \right)$ is a neural network model, and $c_{\text{skip}}(\sigma)$, $c_{\text{out}}(\sigma)$, $c_{\text{in}}(\sigma)$ and $c_{\text{noise}}(\sigma)$ are defined in the paper by Karras et al\cite{NEURIPS2022_a98846e9}.

\begin{figure}[H]
	\centering
	\includegraphics[width=0.7\textwidth]{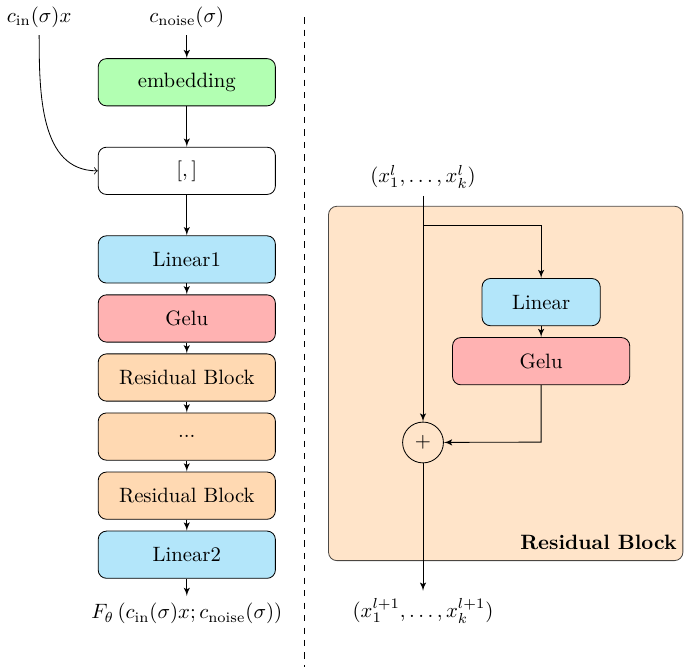}
	\caption{Neural Network structure of $F_{\theta}$ for GMM}
 \label{fig:nngmm}
\end{figure}

We model $F_{\theta} \left( c_{\text{in}}(\sigma)\x;c_{\text{noise}}(\sigma) \right)$ as a  Multi-Layer Perceptron with Residual connections, whose structure is illustrated in Figure \ref{fig:nngmm}.

Here are some more detailed settings for the 1000 dimensional GMM:
\begin{itemize}
	\item Training data is gathered by taking 20000 samples from each Gaussian well.
	\item the embedding layer is a SinusoidalEmbedding layer with emb\_size=80.
	\item The output of the embedding layer is concatenated with $c_{\text{in}}(\sigma)\x$ to form a data vector of size concat\_size.
	\item Linear1 is a fully connected layer with an input dimension of concat\_size and an output dimension of hidden\_size=2000.
	\item The network has 10 Residual Blocks.
        \item Linear is a fully connected layer with both input and output dimensions of hidden\_size=2000.
        \item Linear2 is a fully connected layer with an input dimension of hidden\_size=2000 and an output dimension of output\_size=1000.
\end{itemize}

To integrate the ODE, we use the 2nd order Heun integrator\cite{ascher1998computer}. The time interval for the integration is \([t_{\text{min}}, t_{\text{max}}]\), where \(t_{\text{min}} = 0.01\) and \(t_{\text{max}} = 15\). This interval is discretized into \(N\) subintervals for numerical integration. The boundaries of the \(i\)-th subinterval are defined as follows:
\begin{equation}
t_{i} = \left({t_{\text{min}}}^{\frac{1}{\rho}}+ \frac{i-1}{N-1}\left({t_{\text{max}}}^{\frac{1}{\rho}}- {t_{\text{min}}}^{\frac{1}{\rho}} \right)\right)^{\rho} \,.
\end{equation}
In our implementation, we use \(N = 100\) and \(\rho = 3\).

\subsection{Flow model for Chignolin}
For Chignolin, we use the following functions for \(s(t)\) and \(\sigma(t)\):
\begin{eqnarray}
s(t) & = & \sqrt{\alpha(t)} \\
\sigma(t) & = & \frac{\sqrt{1 - \alpha(t)}}{\sqrt{\alpha(t)}}
\end{eqnarray}
Here, \(\alpha(t)\) is a function that decreases from 1 to 0 as \(t\) increases from 1 to \(T\). At integer time \(t\), \(\alpha(t)\) is defined as:
\begin{equation}
\alpha(t) = \prod_{s=1}^t \left( 1 - \beta_s \right) \,,
\end{equation}
Where \(\beta_s\) is defined as:
\begin{equation}
\beta_s = \text{Sigmoid}(\tau_s) \cdot (\beta_{\text{max}} - \beta_{\text{min}}) + \beta_{\text{min}}
\end{equation}

Here, \(\text{Sigmoid}(x)\) denotes the sigmoid function defined as 
\begin{equation}
	\text{Sigmoid}(x) = \frac{1}{1 + e^{-x}}
\end{equation}	
\(\tau_s\) is an equally spaced time sequence given by
\begin{equation}
\tau_s = -10 + 20 \cdot \frac{s - 1}{T - 1}
\end{equation}
For the diffusion steps over \(T = 1000\) steps, \(\beta_{\text{min}}\) is chosen to be \(1 \times 10^{-5}\), and \(\beta_{\text{max}}\) is chosen to be \(1 \times 10^{-2}\). For non-integer time \(t\), we interpolate between the values of \(\beta(t)\) at integer times to determine its value.

\begin{figure}[H]
	\centering
	\includegraphics[width=0.7\textwidth]{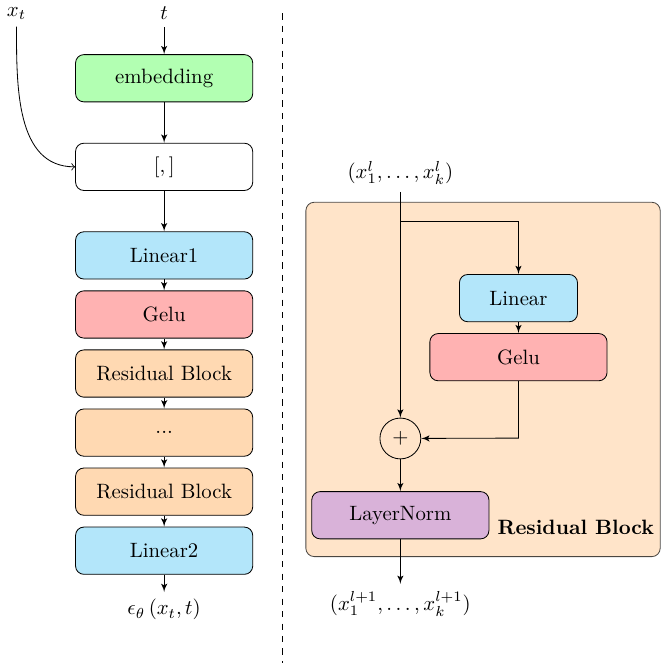}
	\caption{Neural Network structure of $\epsilon_{\theta}$ for Chignolin}
 \label{fig:nncgn}
\end{figure}

The score function is parameterized in the following way:
\begin{equation}
\nabla_{\x} \log p_t(\x)  = -\frac{\epsilon_{\theta}\left( \x_t ,t \right)} {s(t)\sigma(t)} \,,
\end{equation}
where \(\epsilon_{\theta}\left( \x_t ,t \right)\) is neural network whose structure is shown in Figure \ref{fig:nncgn}. As one can see, its only difference with the one used for GMM is that it has an additional LayerNorm layer in the Residual Block.

Here are some more detailed settings:
\begin{itemize}
	\item The training dataset is sourced from Frank Noé's research group's website and requires centering of the data.
	\item The embedding layer is a SinusoidalEmbedding layer with an embedding size of 10.
	\item The output of the embedding layer is concatenated with \(\x_t\) to form data of size concat\_size.
	\item Linear1 is a fully connected layer with an input dimension of concat\_size and an output dimension of hidden\_size = 2048.
	\item The network contains 12 Residual Blocks as hidden layers.
	\item Linear is a fully connected layer with both input and output dimensions of hidden\_size = 2048.
	\item Linear2 is a fully connected layer with an input dimension of hidden\_size = 2048 and an output dimension of output\_size = 525.
\end{itemize}

To integrate the ODE, we use the Heun integrator. The time interval for the integration is \([t_{\text{min}}, t_{\text{max}}]\), where \(t_{\text{min}} = 1\) and \(t_{\text{max}} = 1000\). This interval is discretized into \(N\) subintervals for numerical integration. The boundaries of the \(i\)-th subinterval are defined as follows:
\begin{equation}
t_i = \left({t_{\text{min}}}^{\frac{1}{\rho}}+ \frac{i-1}{N-1}\left({t_{\text{max}}}^{\frac{1}{\rho}}- {t_{\text{min}}}^{\frac{1}{\rho}} \right)\right)^{\rho} \,.
\end{equation}
In our implementation, we use \(N = 100\) and \(\rho = 2\).

\begin{figure}[H]
	\centering
	\includegraphics[width=0.7\textwidth]{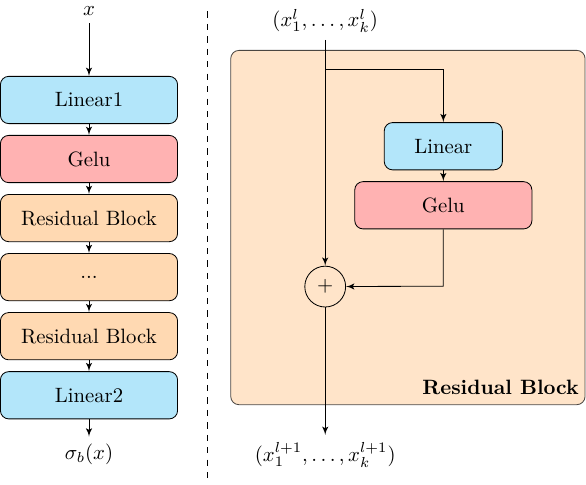}
	\caption{Neural Network structure for \(\sigma_b(\x)\)}
 \label{fig:sigb}
\end{figure}

\section{backward noise scaling function for GMM
	and Chignolin \label{appendix:bnv_used}}

\(\sigma_b(\x)\) is modeled as a neural network whose structure is shown in Figure \ref{fig:sigb}.

Here are some detailed settings:
\begin{itemize}
	\item Linear1 is a fully connected layer with an input dimension of concat\_size=D and an output dimension of hidden\_size=40.
	\item The network contains 10 Residual Blocks as hidden layers.
	\item Linear is a fully connected layer with both input and output dimensions of hidden\_size=40.
	\item Linear2 is a fully connected layer with an input dimension of hidden\_size=40 and an output dimension of output\_size=1.
\end{itemize}

\section{Different choice of forward noise variance \label{appendix:fnv}}
\begin{figure}[H]
	\centering
	\includegraphics[width=1.0\textwidth]{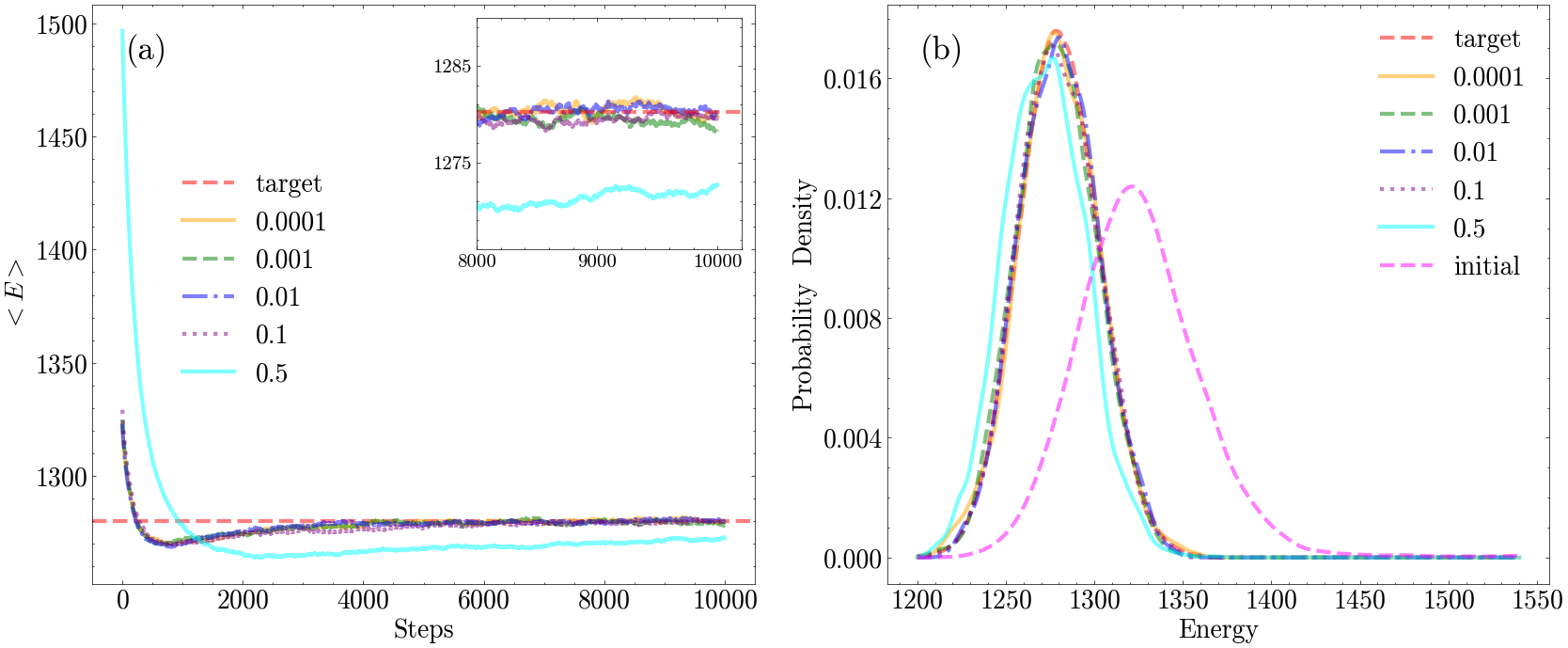}
	\caption{Sampling the Boltzmann distribution  of 1000-dimensional GMM using FP with different choices of \(\sigma_f\)}
 \label{fig:diffsig_gmm}
\end{figure}

For the 1000-dimensional GMM, we experimented with the following values for \(\sigma_f\): \(0.0001, 0.001, 0.01, 0.1, 0.5 \). The results of Metropolis MC with the FP method are summarized in Figure \ref{fig:diffsig_gmm}. From this figure, it is evident that our FP method can accurately reproduce the target distribution when \(\sigma_f\) ranges from \(0.0001\) to \(0.1\). The number of MC steps required for convergence is also similar for these values. Even when \(\sigma_f = 0.5\), the energy distribution obtained only shows a slight leftward shift compared to the target distribution.

\begin{figure}[H]
	\centering
	\includegraphics[width=1.0\textwidth]{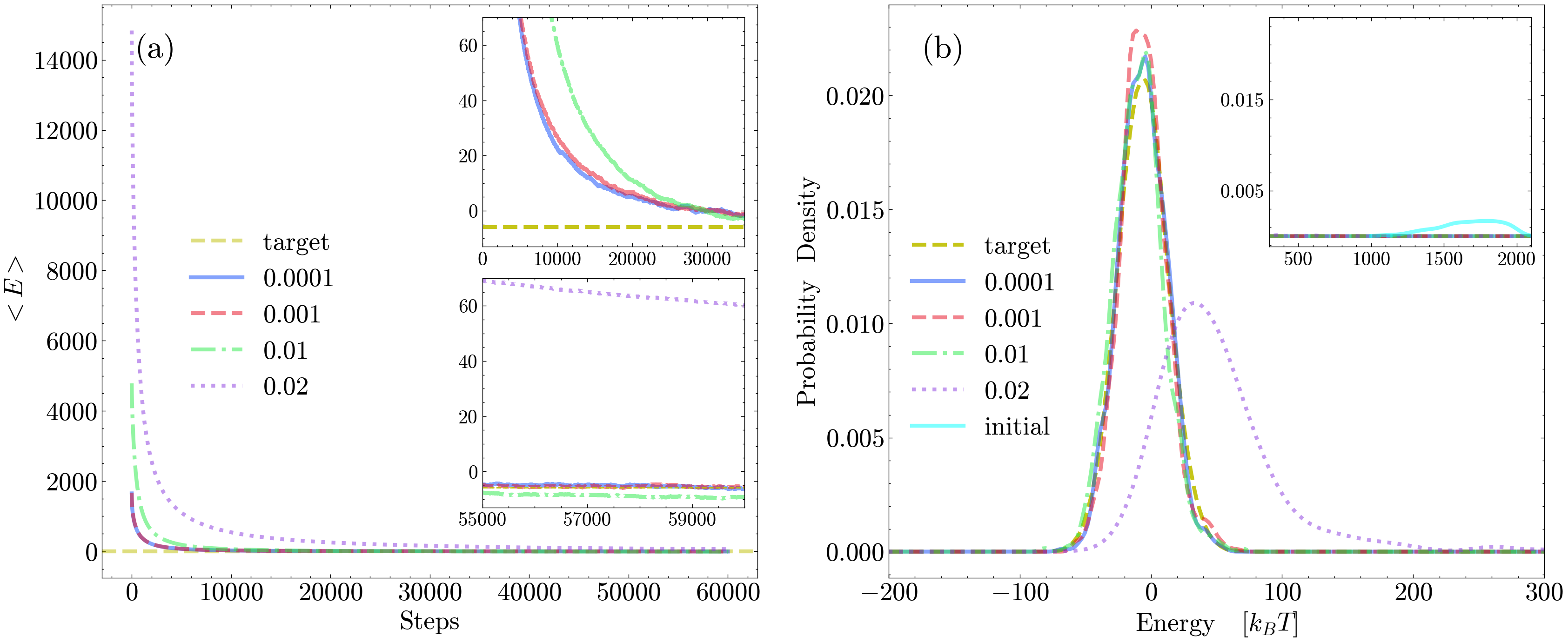}
	\caption{Sampling the Boltzmann distribution of Chignolin using FP with different choices of \(\sigma_f\)}
 \label{fig:diffsig_cgn}
\end{figure}

For the Chignolin protein, we experimented with the following values for \(\sigma_f\): 0.0001 nm, 0.001 nm, 0.01 nm, 0.02 nm and 0.5 nm. The results of the Metropolis MC using the FP method are summarized in Figure \ref{fig:diffsig_cgn}. The FP method accurately reproduces the target distribution when \(\sigma_f\) ranges from 0.0001 nm to 0.001 nm, with a similar number of MC steps required for convergence. When \(\sigma_f = 0.01 \) nm, the energy distribution shows only a very slight leftward shift compared to the target distribution. However, when \(\sigma_f\) is increased to 0.02 nm, the energy distribution begins to deviate significantly from the target distribution.

\section{Different choice of the number of coordinates updated \label{appendix:partial}}
\begin{figure}[H]
	\centering
	\includegraphics[width=1.0\textwidth]{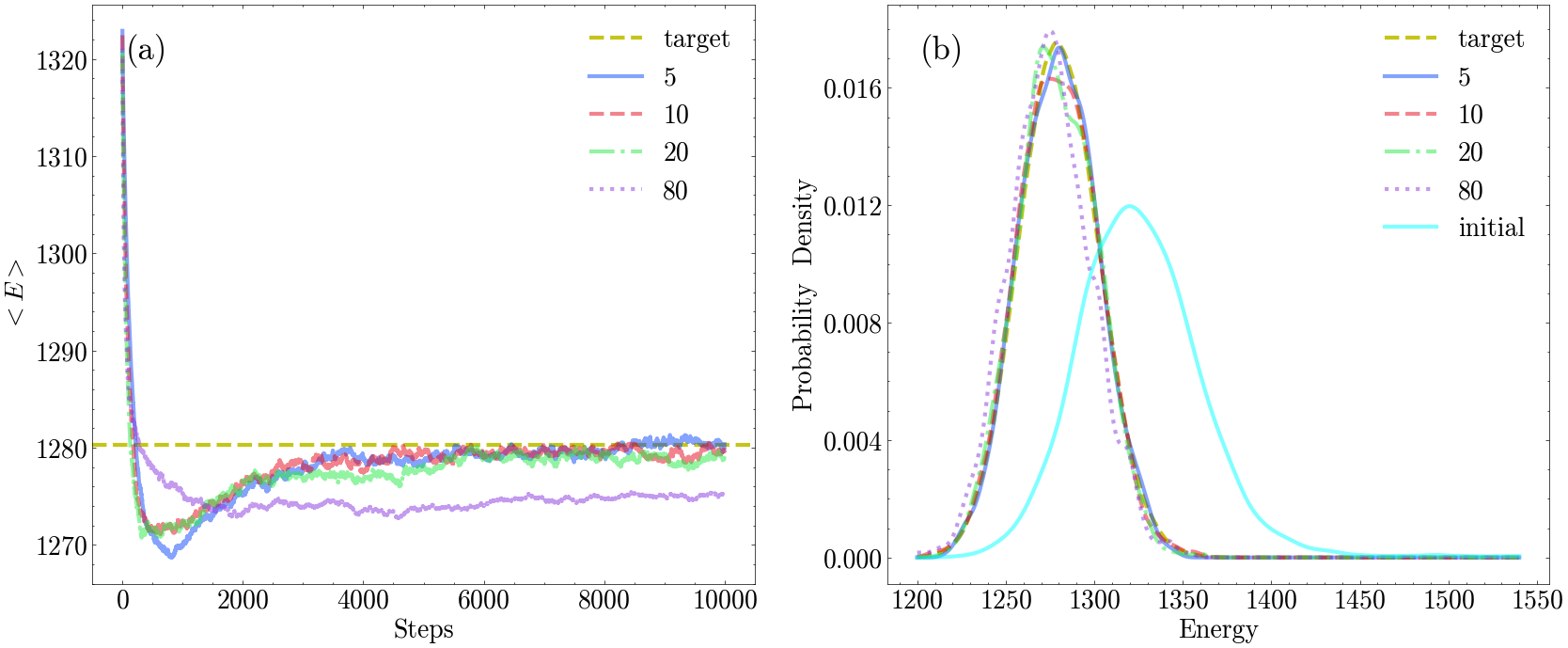}
	\caption{Sampling the Boltzmann distribution of 1000-dimensional GMM using FP with different number of coordinates of $z$ and $\epsilon$ updated per MC step. Legend entries, such as "10," indicate the number of coordinates of 
$z$ and $\epsilon$ that are updated at each step.}
\label{fig:diffK_gmm}
\end{figure}

For the 1000-dimensional GMM, we experimented with updating different numbers of coordinates of the random variables \(\z\) and \(\epsilon\). The results are summarized in Figure \ref{fig:diffK_gmm}. When updating up to 20 coordinates, the energy distribution and the number of MC steps required to converge remained consistent. However, a significant deviation from the target energy distribution was observed when the number of updated coordinates reached 80. This deviation arises because updating too many coordinates at each step significantly reduces the acceptance rate, effectively freezing the MC dynamics.

\begin{figure}[H]
	\centering
	\includegraphics[width=1.0\textwidth]{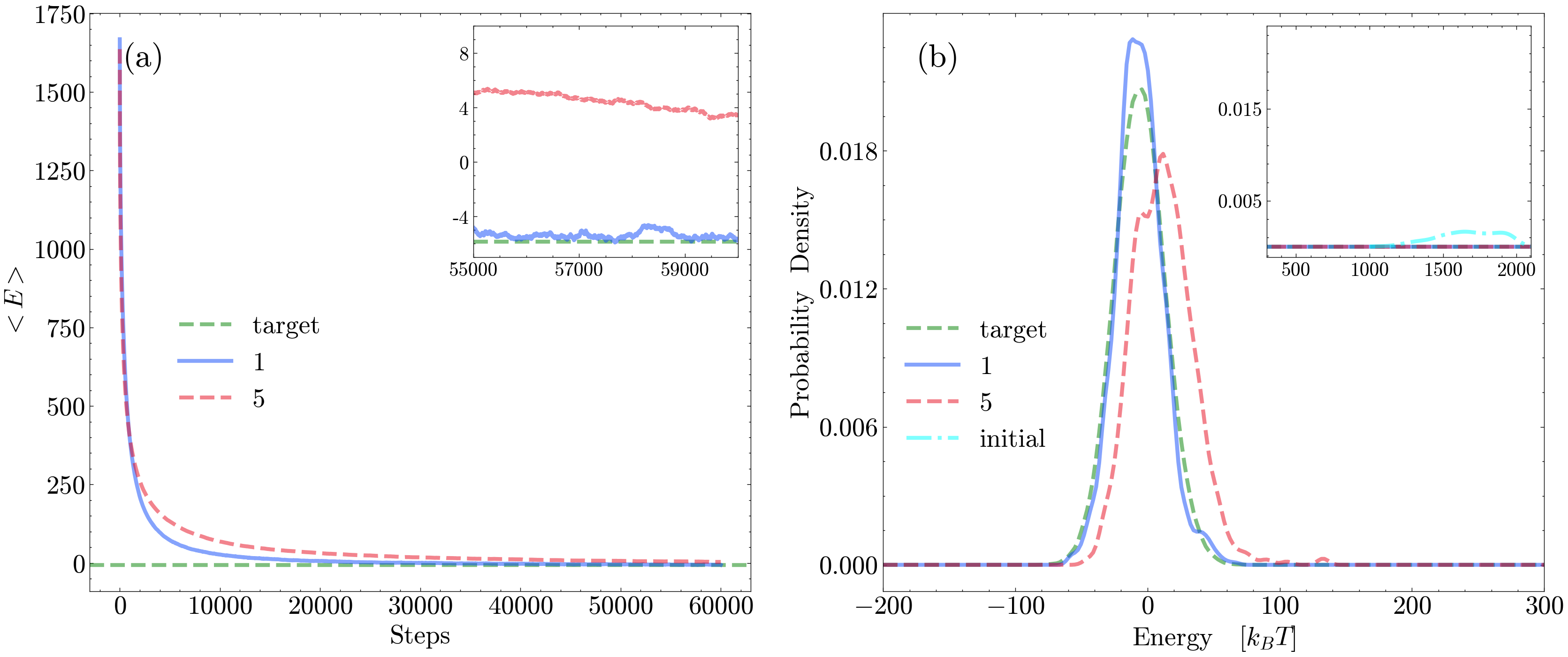}
	\caption{Sampling the Boltzmann distribution of Chignolin using FP with different number of coordinates of $z$ and $\epsilon$ updated per MC step.}
\label{fig:diffK_cgn}
\end{figure}

For the Chignolin protein, the acceptance rate is already very low even when we only update 1 coordinate of \(\z\) and \(\epsilon\) per step.  Increasing the number of updated coordinates to 5 further decreases the acceptance rate, causing the energy distribution obtained to deviate from the target distribution, as shown in Figure \ref{fig:diffK_cgn}. 

\end{document}